\documentclass[11pt]{article}

\usepackage[preprint]{acl}
\usepackage{times}
\usepackage{latexsym}

\usepackage[T1]{fontenc}

\usepackage[utf8]{inputenc}

\usepackage{microtype}

\usepackage{inconsolata}
\usepackage{makecell}

\usepackage{graphicx}
\usepackage{subcaption}
\usepackage{amsmath} 
\usepackage{amssymb}
\usepackage{multirow}
\usepackage{algorithm}
\usepackage{algorithmic}
\usepackage{booktabs}
\usepackage{tabularx}
\usepackage{array}

\usepackage{pifont}
\newcommand{\cmark}{\ding{51}} 
\newcommand{\xmark}{\ding{55}} 

\usepackage{bm} 

\title{Process In-Context Learning: \\Enhancing Mathematical Reasoning via Dynamic Demonstration Insertion}

\author{
    Ang Gao\textsuperscript{1}\thanks{Equal contribution.}
    Changshuo Zhang\textsuperscript{1}\footnotemark[1]
    Xiao Zhang\textsuperscript{1}\thanks{Corresponding author (e-mail: zhangx89@ruc.edu.cn).}
    \textbf{Deyang Li}\textsuperscript{\textbf{2}}\\
    \textbf{Minjun Zhao}\textsuperscript{\textbf{2}}
    \textbf{Fangchao Liu}\textsuperscript{\textbf{2}}
    \textbf{Xinyu Zhang}\textsuperscript{\textbf{2}} \\
    \textsuperscript{1}Gaoling School of Artificial Intelligence, Renmin University of China \\
    \textsuperscript{2}Huawei Poisson Lab, China
}

\begin{document}
\maketitle
\begin{abstract}
In-context learning (ICL) has proven highly effective across diverse large language model (LLM) tasks. However, its potential for enhancing tasks that demand step-by-step logical deduction, such as mathematical reasoning, remains underexplored. A core limitation of existing ICL approaches is their static use of demonstrations: examples are pre-selected before inference and remain fixed, failing to adapt to the dynamic confusion points that often arise during multi-step reasoning such as ambiguous calculations or logical gaps. These unresolved confusion points can lead to cascading errors that degrade final accuracy.  To tackle this issue, we propose Process In-Context Learning (PICL), a dynamic demonstration integration framework designed to boost mathematical reasoning by responding to real-time inference needs. PICL operates in two stages: 1)~it identifies potential confusion points by analyzing semantics and entropy in the reasoning process and summarizes their core characteristics; 2)~upon encountering these points, it retrieves relevant demonstrations from the demonstration pool that match the confusion context and inserts them directly into the ongoing reasoning process to guide subsequent steps. Experiments show that PICL outperforms baseline methods by mitigating mid-inference confusion, highlighting the value of adaptive demonstration insertion in complex mathematical reasoning.
\end{abstract}

\section{Introduction}

Mathematical reasoning, a cornerstone of AI intelligence assessment, requires step-by-step logical deduction and remains challenging for even state-of-the-art models~\cite{zhang2019gap, liu2024makes}. In-context learning (ICL) has emerged as a pivotal paradigm to address this, enabling models to improve reasoning via task-related demonstrations without parameter updates~\cite{dong2022survey}. However, recent studies have revealed that ICL’s effectiveness in mathematical reasoning is not guaranteed. Simply prepending a set of demonstrations before inference can even yield negative effects, especially in multi-step logical tasks~\cite{liu2024makes, zheng2025curse, cheng2025revisiting}.

\begin{figure}[t] 
    \centering
    \begin{subfigure}[b]{0.45\textwidth} 
        \centering
        \includegraphics[width=\textwidth]{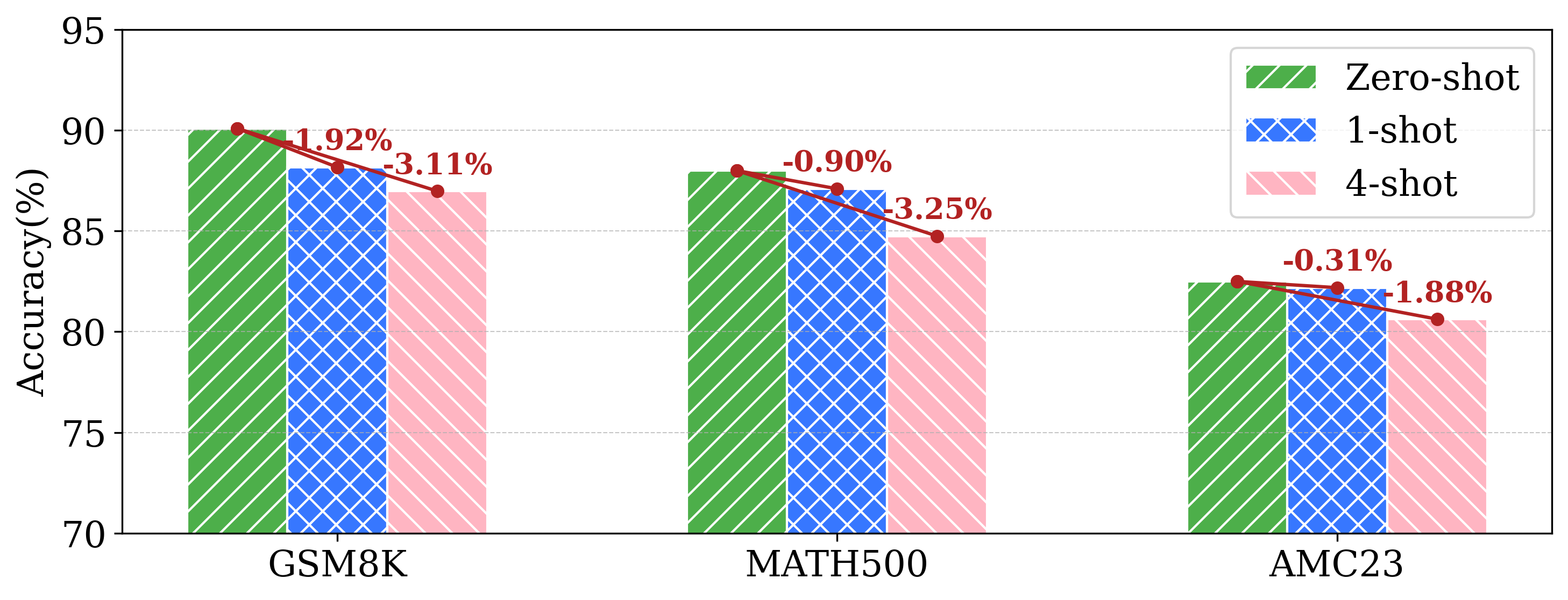} 
    \vspace{-0.6cm} 
        \caption{Accuracy comparison of zero-shot and few-shot.}
        \label{fig:sub1} 
    \end{subfigure}

    \begin{subfigure}[b]{0.45\textwidth} 
        \centering
        \includegraphics[width=\textwidth]{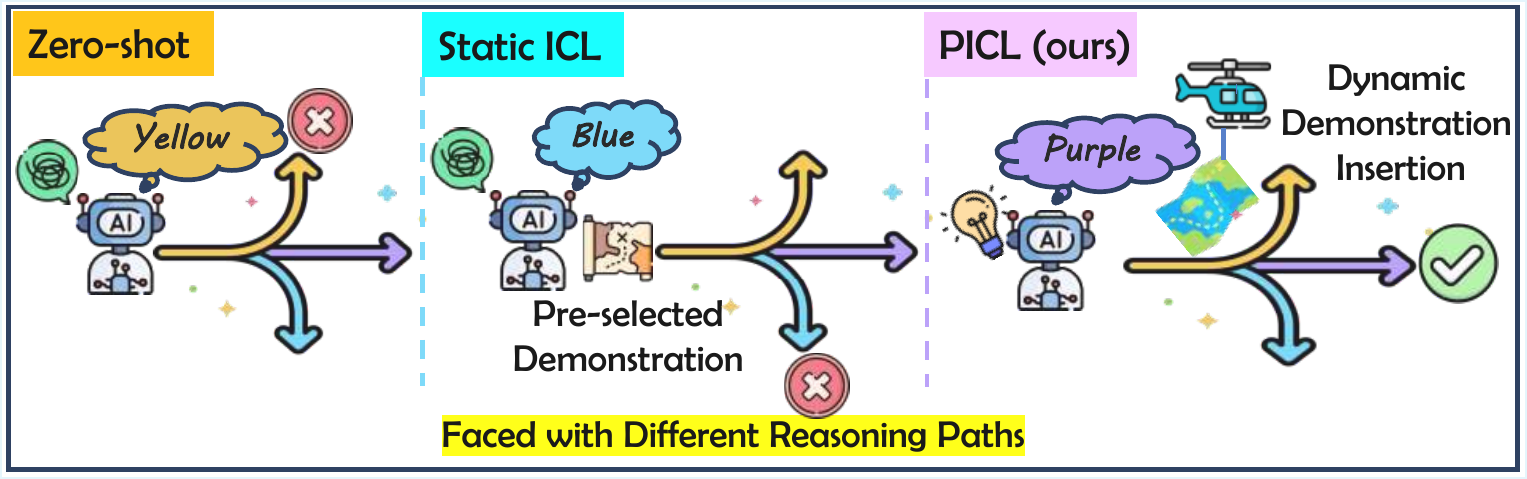} 
        \caption{Illustration of different inference methods.}
        \label{fig:sub2} 
    \end{subfigure}

    \caption{Performance of static ICL and the conceptual framework of PICL. (a) Performance comparison of Deepseek-R1-Distilled-Qwen-7B under zero-shot, 1-shot and 4-shot ICL settings on three math benchmarks. (b) A schematic illustration comparing three inference paradigms when encountering a reasoning path selection challenge. Left: Zero-shot prompting. Middle: Conventional static few-shot ICL. Right: Our proposed PICL framework, which inserts targeted demonstrations to provide adaptive guidance.} 
    \label{fig:intro} 
\end{figure}

Existing ICL methods pre-select demonstrations upfront and fix them throughout inference~\cite{wies2023learnability, min2022rethinking}, but such static integration fails to align with the dynamic nature of mathematical reasoning. In multi-step deduction, models often hit ``confusion points'' such as hesitating between calculation paths or misidentifying logical transitions~\cite{tang2025refcritic}. Static demonstrations not only fail to resolve these but may exacerbate errors by introducing irrelevant reasoning patterns, leading to cascading mistakes in subsequent steps~\cite{liu2024makes}. Worse, even advanced demonstration selection strategies such as focusing on similarity or influence overlook in-process reasoning dynamics, making them ineffective or counterproductive in complex mathematical tasks~\cite{cheng2025revisiting, zheng2025curse}. Our experiments reinforce this observation: static ICL approaches offer limited benefits for mathematical reasoning and lead to performance degradation in many cases. 
As shown in Fig.~\ref{fig:sub1}, static few-shot demonstrations fail to outperform the zero-shot baseline, and simply increasing their number can further degrade performance, underscoring the fundamental limitations of this fixed approach. This highlights an urgent need. Instead of fixed pre-inference demonstrations, we must inject targeted examples during the reasoning process to provide adaptive guidance when models need it most.

To bridge this gap, we propose PICL (Process In-Context Learning), a novel method whose advantages over static methods are illustrated in Fig.~\ref{fig:sub2}. This figure depicts a common failure mode in multi-step reasoning: a ``confusion point'' where the model must select the correct logical path. As shown, traditional approaches such as zero-shot learning and static ICL struggle to handle this scenario: the former lacks guidance entirely, while the latter’s fixed demonstrations, which lack targeting, prove unhelpful. Both ultimately lead the model down an erroneous reasoning path. In contrast, PICL operates differently: it actively detects points of uncertainty in the reasoning process and dynamically intervenes by injecting highly relevant examples. This adaptive demonstration insertion provides the precise guidance needed to resolve confusion, redirecting the model back to the correct reasoning path.

PICL consists of two core stages, tightly coupled with the inference process:
First, the \emph{Confusion Point Detection and Summarization} stage: During multi-step reasoning, PICL detects the model’s intermediate outputs and identifies ``confusion points'' using semantics and entropy indicators of prediction uncertainty~\cite{wang2025beyond}. It then summarizes the core characteristics of these confusion points (such as ambiguities in algebraic transformation or misalignments of logical premises) to determine the specific guidance required.
Second, the \emph{Confusion-Based Demonstration Selection \& Insertion} stage: Based on the summarized characteristics of the confusion points, PICL retrieves the most relevant demonstrations from a pre-constructed example pool. These demonstrations are designed to address the identified issues (such as resolving the same algebraic ambiguity) and are inserted directly into the ongoing reasoning process. This provides targeted guidance to correct the model’s reasoning trajectory and alleviate confusion. In summary, our work makes three key contributions:


\begin{itemize}
    \item 
\textbf{A novel dynamic demonstration insertion paradigm:} We move beyond static ICL by introducing process-aware demonstration adaptation, enabling real-time guidance tailored to the model’s in-process confusion point and addressing the rigidity of existing methods.
    \item 
\textbf{A confusion point detection mechanism:} We propose using semantics and entropy to detect reasoning uncertainty and summarize confusion characteristics, providing a concrete basis for targeted demonstration retrieval.
    \item 
\textbf{Empirical validation across benchmarks:} Experiments on mathematical reasoning datasets show PICL outperforms state-of-the-art static ICL methods, with significant gains in multi-step tasks where confusion points are most prevalent.
\end{itemize}

\begin{figure*}[t] 
    \centering 
    \includegraphics[width=0.95\textwidth]{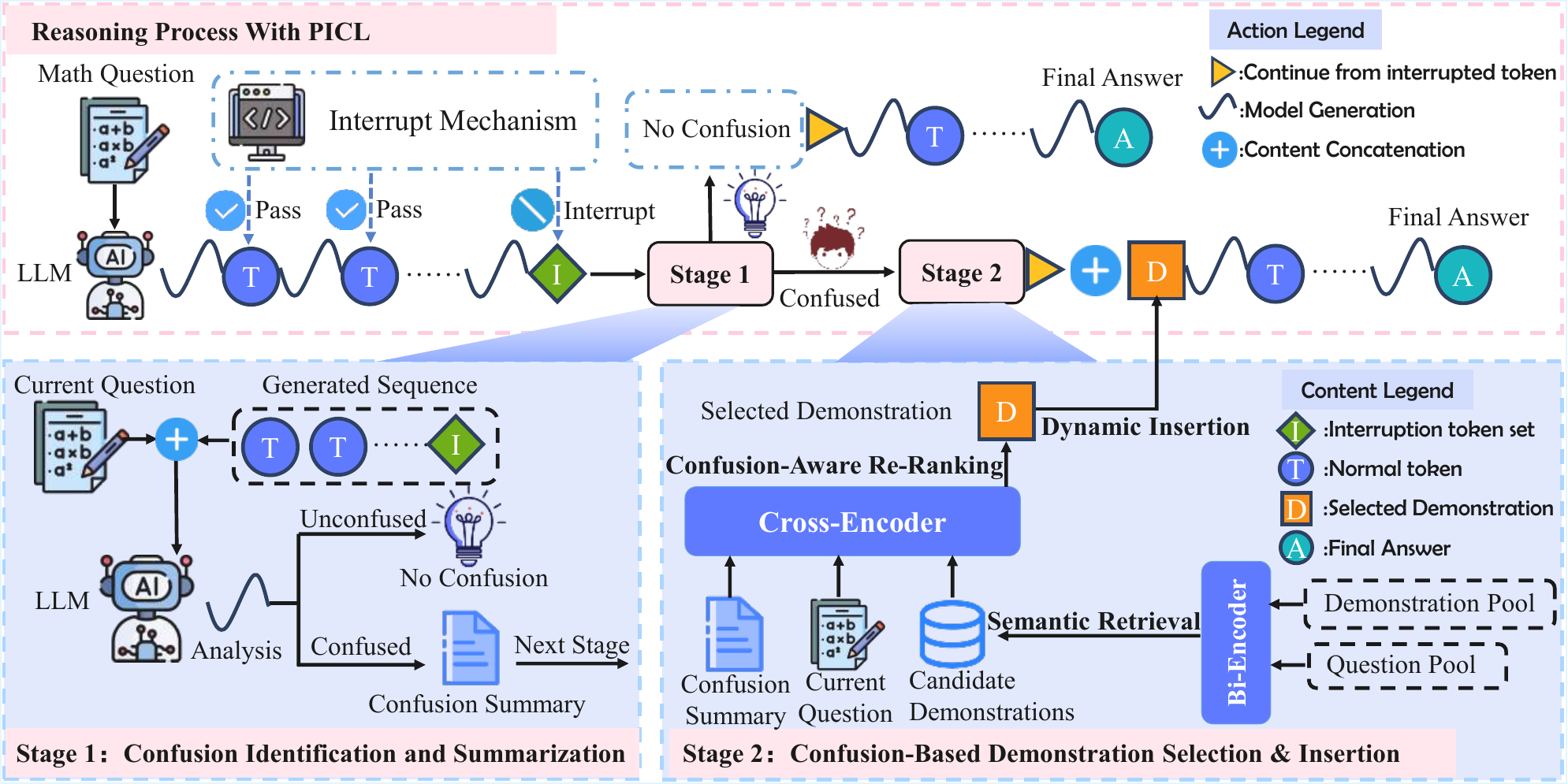} 
    \caption{Process In-Context Learning (PICL) for enhancing mathematical reasoning of language models. It features two stages: an Interrupt Mechanism detects confusion via semantics and entropy in reasoning. Upon detection, relevant demonstrations are dynamically retrieved from a pool and inserted, refining reasoning process.} 
    \label{fig:main} 
\end{figure*}

\section{Related Work}
\paragraph{In-Context Learning.}In-context learning (ICL) enables LLMs to perform tasks without parameter updates by conditioning on demonstrations~\cite{brown2020language}. Prior work studies the impact of scale, demonstration format, and data distribution~\cite{dong2022survey, chen2021evaluating, min2022rethinking, wei2022chain}. Demonstration selection is critical for ICL: unsupervised methods use criteria like similarity, diversity and uncertainty~\cite{liu-etal-2022-makes, levy2022diverse, wu2022self, nguyen2023context}, while supervised methods rely on learning-based policies~\cite{rubin2021learning, ye2023compositional, zhang2022active, wang2023learning}. However, most methods adopt a static paradigm, using the same exemplars throughout the process, which can limit adaptability in complex reasoning.

\paragraph{Mathematical Reasoning Enhancement.}Math reasoning is difficult for LLMs due to multi-step inference and numerical precision. Recent methods improve performance via prompting and augmentation. Structured prompting uses Chain-of-Thought to guide reasoning~\cite{yao2023tree}. Code-based method alleviate errors by prompting code generations~\cite{gao2023pal, chen2021evaluating}. Others augment models with retrieval, web search,  tools~\cite{lewis2020retrieval, nakano2021webgpt, schick2023toolformer, webthinker}, and math-focused fine-tuning with process supervision~\cite{yu2023metamath, lewkowycz2022solving, cobbe2021training, lightman2023let}. Our work targets ICL for better mathematical reasoning support.

\paragraph{ICL for Mathematical Reasoning.}Most mathematical ICL methods remains static: exemplars are chosen beforehand using criteria such as influence functions~\cite{guo2024makes}, mutual information~\cite{sorensen2022information}, perplexity/entropy~\cite{gonen2022demystifying, lu2021fantastically}, or semantic similarity via BM25/BERT/BGE-M3~\cite{robertson2009probabilistic, devlin2019bert, chen2024bge}. However, recent evidence suggests static ICL can be ineffective or even harmful for math reasoning~\cite{cheng2025revisiting, zheng2025curse}. Motivated by these limitations, we propose PICL, a dynamic framework that inserts demonstrations at uncertainty moments during reasoning, enabling more targeted and adaptive problem solving.


\section{Problem Formulation}
\label{sec:preliminary}

In-context learning (ICL) is a powerful paradigm that helps LLMs solve new problems by conditioning on demonstrations. This avoids the need for parameter fine-tuning.

The conventional approach, which we term static ICL, operates in a single, fixed pass. Given a new query $q$ and a demonstration pool $\mathcal{D} = \{d_1, \dots, d_N\}$, a set of $k$ demonstrations is selected before the model begins its reasoning process. Each demonstration $d_i$ typically consists of an example problem $x_i$ and its corresponding solution or reasoning process $y_i$, i.e., $d_i = (x_i, y_i)$. This selection is usually based on the metrics between the new query $q$ and the problem component $x_i$ of each demonstration. These selected demonstrations, $\{d_{s_1}, \dots, d_{s_k}\}$, are then prepended to the query to form a single, static prompt:
\[
\text{Prompt}_{\text{static}} = d_{s_1} \circ d_{s_2} \circ \dots \circ d_{s_k} \circ q.
\]
Here, $\circ$ denotes the concatenation operator that sequentially appends demonstrations and the query to form a single prompt.
The model $M$ conditions on this entire prompt to generate the answer in one uninterrupted sequence.

The defining limitation of this static paradigm is its rigidity, which is especially acute in tasks requiring complex, multi-step reasoning such as mathematics. This motivates our work on PICL, a dynamic framework that operates on the same demonstration pool $\mathcal{D}$. However, instead of statically pre-selected demonstrations, PICL dynamically inserts demonstrations $\{d_{s_1}, \dots, d_{s_k}\}$ into the model's reasoning process precisely when and where confusion is detected.

\section{PICL: The Proposed Method}
We propose Process In-Context Learning (PICL), a novel framework designed to enhance the mathematical reasoning capabilities of language models by dynamically inserting demonstrations during the inference process. 
\subsection{Method Overview}
As illustrated in Fig.~\ref{fig:main}, PICL operates in two primary stages: 
\emph{Stage~1: Confusion Point Identification and Summarization},  where it identifies potential confusion points by analyzing semantics and entropy during the reasoning process and summarizes the context of this confusion; 
and \emph{Stage~2: Confusion-Based Demonstration Selection \& Insertion}, where upon encountering these points, PICL retrieves relevant demonstrations from a pre-constructed demonstration pool, tailored to address the specific confusion, and inserts them into the reasoning process to improve performance on mathematical tasks.
We detail these two stages in the following subsections.

\subsection{Confusion Identification and Summarization}
The first stage of PICL is to detect moments of uncertainty in the model's reasoning process. This is achieved by identifying confusion points, which are instances where the model exhibits signals of hesitation or deviates from a confident reasoning trajectory.

Our detection starts from an Interrupt Mechanism, which is triggered by a predefined set of interruption tokens, denoted as $\mathcal{V}_{\text{int}}$. The selection of this vocabulary is based on a dual criterion. First, from a \textbf{semantic} perspective, we deliberately choose tokens that linguistically signal reflection, transition, or hesitation. Words such as ``wait'' or ``maybe" are not arbitrary; they explicitly mark moments where a reasoning process is being re-evaluated or redirected. By monitoring for these specific tokens, we target junctures that are most likely to represent complex reasoning steps or potential points of confusion. 

Second, this semantic intuition is empirically substantiated by a key statistical property: these chosen tokens consistently exhibit high \textbf{entropy}. At each step $t$ when the model is about to generate the next token, it first computes a logit vector $\bm{l}_t \in \mathbb{R}^{|\mathcal{V}|}$ over its vocabulary $\mathcal{V}$. This vector is then transformed into a probability distribution $\bm{p}_t$ via the softmax function. The model's local uncertainty at this specific step is then quantified by the Shannon entropy of this probability distribution:
\[
H(\bm{p}_t) = -\sum_{v \in \mathcal{V}} p_t(v) \log p_t(v).
\]
A high value of $H(\bm{p}_t)$ indicates that the model's prediction is diffuse across many tokens, indicating greater uncertainty.

When the model generates a token $y_t \in \mathcal{V}_{int}$, the generation process is temporarily paused. Following this interruption, PICL initiates a self-reflection step to determine if a meaningful confusion exists. The model is prompted to analyze its preceding output $Y_{<t}$ in the context of the query $q$. If and only if a specific point of confusion is identified, the model is tasked with generating a concise summary of the issue. We formalize this process as:
\[
\mathcal{C} = M(\text{Prompt}_{\text{detect}}, q, Y_{<t}),
\]
where $M$ is the large language model itself, and $\text{Prompt}_{\text{detect}}$ is a specialized instruction that guides the model to articulate its confusion. The resulting confusion summary, $\mathcal{C}$, might describe issues like ``ambiguity in applying a specific formula'' or ``uncertainty about the next logical step.'' If no confusion is found, $\mathcal{C}$ is considered empty, and no further action is taken. This step identifies the points of confusion in the model's reasoning process, specifies the locations for dynamic demonstration insertion, and generates a confusion summary to facilitate the selection of demonstrations.


\subsection{Confusion-Based Demonstration Selection \& Insertion}
Once a confusion summary $\mathcal{C}$ is generated, PICL initiates a three-phase process to select the most relevant demonstrations to resolve the identified issue. This process leverages the query, the confusion summary, and the candidate demonstrations.

\textbf{Phase 1: Semantic Retrieval.}
Given the current question \( q \) and a demonstration pool \( \mathcal{D} = \{d_1, \dots, d_N\} \), the initial stage aims to efficiently narrow down the vast demonstration pool $\mathcal{D}$ to a smaller set of semantically relevant candidate demonstrations. We employ a computationally efficient Bi-Encoder architecture for this task. The problem query \( q \) and each demonstration $d_i \in \mathcal{D}$ are independently encoded into dense vector representations using a pre-trained sentence transformer like BERT:
\[
\bm{h}_q = f_{\text{bi-enc}}(q), \quad \bm{h}_{d_i} = f_{\text{bi-enc}}(d_i).
\]
We then compute the cosine similarity between the query vector $\bm{h}_q$ and each demonstration vector $\bm{h}_{d_i}$: 
\[
\text{sim}(q, d_i) = \frac{\bm{h}_q \cdot \bm{h}_{d_i}}{\|\bm{h}_q\| \|\bm{h}_{d_i}\|}.
\]
The top-$N$ demonstrations with the highest similarity scores are selected to form a candidate set \(\mathcal{D_N}\), reducing the search space for the next stage.

\begin{algorithm}[tb]
\caption{Process In-Context Learning (PICL)}
\label{alg:picl}

\begingroup
\small

\textbf{Input}: Large Language model $M$, query $q$, demonstration pool $\mathcal{D}$.\\
\textbf{Parameter}: Interruption tokens $\mathcal{V}_{\text{int}}$, insertion count $k$, max interventions $r$, retrieval candidates $N$.\\
\textbf{Output}: The final answer $A$.\\
\begin{algorithmic}[1] 
\STATE Initialize $Y \leftarrow \emptyset$, $intervention\_cnt \leftarrow 0$
\WHILE{\textbf{not} \texttt{IsEndOfSequence}($Y$) \textbf{and} $intervention\_cnt < r$}
    \STATE $y_{\text{next}} \leftarrow \texttt{GenerateNextToken}(M, q, Y)$
    \IF{$y_{\text{next}} \in \mathcal{V}_{\text{int}}$}
        \STATE $intervention\_cnt \leftarrow intervention\_cnt + 1$
        \STATE $\mathcal{C} \leftarrow \texttt{GenerateConfusionSummary}(M, q, Y)$
        \IF{$\mathcal{C}$ is not empty}  
            \STATE $\mathcal{D}_N \leftarrow \texttt{RetrieveCandidates}(\mathcal{D}, q, N)$
            \STATE $\mathcal{D}_{\text{top-k}} \leftarrow \texttt{RerankCandidates}(\mathcal{D}_N, q, \mathcal{C}, k)$
            \STATE $Y \leftarrow Y \oplus \mathcal{D}_{\text{top-k}}$
            \COMMENT{Insert demonstrations}
        \ELSE  
            \STATE $Y \leftarrow Y \oplus y_{\text{next}}$
            \\ \COMMENT{No confusion, continue normally}
        \ENDIF
    \ELSE
        \STATE $Y \leftarrow Y \oplus y_{\text{next}}$
    \ENDIF
\ENDWHILE

\WHILE{\textbf{not} \texttt{IsEndOfSequence}($Y$)}
    \STATE $y_{\text{next}} \leftarrow \texttt{GenerateNextToken}(M, q, Y)$
    \STATE $Y \leftarrow Y \oplus y_{\text{next}}$
\ENDWHILE 
\STATE $A \leftarrow \texttt{ExtractAnswerFrom}(Y)$
\STATE \textbf{return} $A$
\end{algorithmic}
\endgroup
\end{algorithm}

\textbf{Phase 2: Confusion-Aware Re-Ranking.}
The second stage performs a fine-grained re-ranking of the candidate demonstrations in \(\mathcal{D_N}\), this time explicitly incorporating the confusion summary $\mathcal{C}$. We utilize a more powerful cross-encoder model, specifically the BGEM3-Reranker~\cite{bgereranker}, which can capture deeper contextual interactions. For each candidate demonstration $d_i \in \mathcal{D_N}$, we construct a composite input by concatenating the original query $q$, the confusion summary $\mathcal{C}$, and the demonstration $d_i$:
\[
\text{Input}_i = \texttt{[CLS]} \, q \, \texttt{[SEP]} \, d_i \, \texttt{[SEP]} \, C \, \texttt{[SEP]}.
\]
This combined input is fed into the cross-encoder, which outputs a relevance score $s_i$ that quantifies the relevance of the demonstration $d_i$ with respect to the query $q$ and the confusion $\mathcal{C}$:
\[
s_i =  f_{\text{cross-enc}}(\text{Input}_i).
\]
The candidate demonstrations are then re-ranked based on these scores in descending order.

\textbf{Phase 3: Demonstration Insertion. } Following the confusion-aware re-ranking, the final step is to dynamically augment the model's reasoning context. From the re-ranked list, we select the top-$k$ demonstrations, where $k$ is a tunable hyperparameter that controls the amount of guidance inserted at each intervention.

These top-$k$ selected demonstrations, denoted as ${d_{s_1}, \dots, d_{s_k}}$, are then injected directly into the inference stream at the point of interruption. This alters the context for subsequent token generation. The modified generation path can be visualized as:
$$
q \rightarrow y_1, \dots, y_{t-1}, \underbrace{d_{s_1}, \dots, d_{s_k}}_{\text{inserted}}, y_t,
$$
where $y_1, \dots, y_{t-1}$ represents the sequence generated before the pause, and $y_t$ is the next token generated by the model, now conditioned on the newly provided examples. This targeted intervention provides immediate guidance to help the model navigate the specific point of confusion and correct its reasoning trajectory.

To ensure the process remains computationally efficient and does not excessively disrupt the reasoning flow, we introduce a crucial control mechanism. We define a hyperparameter, $r$, which sets the maximum number of times the confusion detection and insertion cycle can be triggered for a single problem-solving instance. This cap prevents the model from entering protracted loops of self-correction and ensures that PICL provides targeted assistance without incurring prohibitive computational overhead. This entire iterative process of confusion detection, retrieval, re-ranking, and insertion, which can be repeated up to $r$ times, is formally detailed in Algorithm 1. We also provide a case study in Appendix~\ref{sec:case_study} to illustrate PICL.

\begin{table*}[t]
\centering
\small
\setlength{\tabcolsep}{3.4pt}        
\renewcommand{\arraystretch}{1.06}   

\begin{tabular}{
@{} >{\raggedright\arraybackslash}p{2.35cm}
ccccc
@{\hspace{10pt}}
ccccc @{}}
\toprule
\multirow{2}{*}{Method}
& \multicolumn{5}{c}{\textbf{\itshape Deepseek-R1-Distilled-Qwen-7B}} 
& \multicolumn{5}{c}{\textbf{\itshape Deepseek-R1-Distilled-Llama-8B}} \\
\cmidrule(lr){2-6}\cmidrule(lr){7-11}
& GSM8K & MATH500 & AMC23 & AIME24 & Avg.
& GSM8K & MATH500 & AMC23 & AIME24 & Avg. \\
\midrule

Zero-shot       
& 90.1 & 88.0 & 82.5 & {\underline{50.0}} & {\underline{77.7}}
& 75.5 & 66.4 & 82.5 & {\underline{46.7}} & 67.8 \\

Random          
& 89.8 & 87.4 & 82.5 & 46.7 & 76.6
& 73.6 & 66.0 & 72.5 & 36.7 & 62.2 \\

Best-validate    
& 80.5 & {\underline{88.4}} & 72.5 & 46.7 & 72.0
& 73.4 & 65.2 & 82.5 & 40.0 & 65.3 \\

Similarity       
& 88.4 & 87.4 & 77.5 & 40.0 & 73.3
& {\underline{76.0}} & 65.4 & 82.5 & 43.3 & 66.8 \\

BM25             
& 89.7 & 85.0 & {\underline{87.5}} & 46.7 & 77.2
& 75.2 & 67.2 & 72.5 & 33.3 & 62.1 \\

BGEM3            
& 88.3 & 87.0 & 85.0 & {\underline{50.0}} & 77.6
& 73.7 & 67.6 & 75.0 & {\underline{46.7}} & 65.8 \\

Perplexity       
& 89.4 & 86.2 & {\underline{87.5}} & 40.0 & 75.8
& 73.4 & 67.4 & 80.0 & 36.7 & 64.4 \\

Influence        
& 88.5 & {\underline{88.4}} & 82.5 & 40.0 & 74.8
& 74.9 & {\underline{68.0}} & 80.0 & {\underline{46.7}} & 67.4 \\

LMS3             
& {\underline{90.8}} & 87.0 & 82.5 & 43.3 & 75.9
& 75.4 & 66.6 & {\underline{85.0}} & {\underline{46.7}} & {\underline{68.4}} \\

\textbf{PICL (Ours)}     
& \textbf{91.7} & \textbf{90.2} & \textbf{90.0} & \textbf{60.0} & \textbf{83.0}
&\textbf{76.9}  & \textbf{70.6} & \textbf{90.0} & \textbf{53.3} & \textbf{72.7} \\

\bottomrule
\end{tabular}

\caption{Accuracy (\%) comparison on four datasets across different backbones.
Avg. is computed over GSM8K, MATH500, AMC23, and AIME24. All few-shot baselines utilize one-shot reasoning.
For PICL, parameters are set to \(k=1\) and \(r=1\). The best and runner-up methods are highlighted in bold and underlined, respectively.}
\label{tab:main_result}
\end{table*}

\begin{table}[t]
\centering
\setlength{\tabcolsep}{3.5pt} 
\renewcommand{\arraystretch}{0.92} 
\footnotesize 

\begin{tabular}{cc|cccc}
\hline
\multirow{2}{*}{Stage 1} & \multirow{2}{*}{Stage 2} &
\multirow{2}{*}{GSM8K} & \multirow{2}{*}{MATH500} &
\multirow{2}{*}{AMC23} & \multirow{2}{*}{AIME24} \\
& & & & & \\
\hline
\multicolumn{6}{c}{\itshape Deepseek-R1-Distilled-Qwen-7B} \\
\hline
\xmark  & \cmark  & 91.1 & 89.0 & 82.5 & 53.3 \\
\cmark  & \xmark  & 90.9 & 87.8 & 82.5 & 50.0 \\
\cmark  & \cmark  & \textbf{91.7} & \textbf{90.2} & \textbf{90.0} & \textbf{60.0} \\
\hline
\multicolumn{6}{c}{\itshape Deepseek-R1-Distilled-Llama-8B} \\
\hline
\xmark  & \cmark  & 75.8 & 64.2 & 82.5 & 43.3 \\
\cmark  & \xmark  & 76.6 & 64.6 & 80.0 & 50.0 \\
\cmark  & \cmark  & \textbf{76.9} & \textbf{70.6} & \textbf{90.0} & \textbf{53.3} \\
\hline
\end{tabular}

\caption{Ablation studies of two core stages in PICL.}
\label{tab:abalation_study}
\end{table}

\section{Experiments}
This section evaluates the performance of PICL and other ICL approaches across multiple mathematical datasets. 

\subsection{Experimental Settings}

\subsubsection{Datasets.}We conducted experiments on four mainstream datasets covering problems of varying difficulty levels and types. The selected datasets include GSM8K~\cite{gsm8k}, MATH500~\cite{math500}, AMC23~\cite{AMC23} and AIME24~\cite{AIME}, which span elementary to competition-level difficulty. We also conducted experiments on multiple-choice benchmarks for generation, which are reported in Appendix~\ref{sec:mc_generalization}.


\subsubsection{Models.}We adopt the open-source DeepSeek-R1 reasoning model series as the backbone. The experiments are carried out on Deepseek-R1-Distilled-Qwen-7B and Deepseek-R1-Distilled-Llama-8B~\cite{deepseek}. 


\subsubsection{Baselines.}We compare our approach with traditional ICL demonstration selection methods and select several representative methods as baselines. These methods cover mainstream selection criteria such as similarity, perplexity, influence and stability, described as follows:
\begin{itemize}
    \item \textbf{Zero-shot} makes predictions directly from the input without any demonstrations.
    \item \textbf{Random} serves as a baseline by randomly selecting the demonstrations.
    \item \textbf{Best-validate} selects the demonstration with the highest accuracy, which is evaluated on a validation set.
    \item \textbf{BM25}~\cite{robertson2009probabilistic} uses BM25 retrieval model to select demostrations.
    \item \textbf{Similarity}~\cite{liu2021makes} selects demonstrations based on their semantic similarity to the current query.
    \item \textbf{BGEM3}~\cite{chen2024bge} uses a unified embedding space integrated by multiple information retrieval capabilities.
\item \textbf{Perplexity}~\cite{gonen2022demystifying} selects demonstrations based on the calculated perplexity of each demonstration.
    \item \textbf{Influence}~\cite{nguyen2023context} determines demonstrations by comparing validation accuracy differences between subsets with and without the demonstration.
    \item \textbf{LMS3}~\cite{liu2024makes} balances demonstrations' semantic similarity and inference stability, incorporated with a rejection mechanism.
\end{itemize}
\subsubsection{Evaluation.}  For evaluation, we employ a sampling strategy with a temperature of 0.7 and top\_p of 0.8. We use problem-solving accuracy as evaluation metric.
\subsubsection{Implementation Details.} A crucial aspect of our experimental design is the number of demonstrations provided. We evaluate the average performance of various few-shot methods with varying numbers of shots, comparing them to the zero-shot baseline. As shown in Fig.~\ref{fig:sub1}, it reveals that for traditional static ICL methods, increasing the number of examples from one-shot to four-shot leads to a decrease in average accuracy. Therefore, all traditional ICL methods are evaluated in a one-shot setting. For fairness and a direct comparison, our PICL method is also configured to add at most one example, with parameters set to $r=1$ and $k=1$. This means PICL performs only a single round of confusion detection and selects at most one demonstration, making it directly comparable to the one-shot baseline setup.


\subsection{Main Experiment}
\subsubsection{Main Results.}
The results in Table \ref{tab:main_result} demonstrate the consistent superiority of PICL across different model architectures. On \textit{R1-Qwen-7B} and \textit{R1-Llama-8B}, PICL achieves average accuracies of 83.0\% and 72.7\%, surpassing the strongest respective baselines by significant margins of 5.3 and 4.3 percentage points.

A phenomenon observed across both backbones is the instability of conventional static ICL methods. For current reasoning models, static few-shot baselines frequently underperform the simple Zero-shot setting. This trend is particularly pronounced on the \textit{R1-Qwen-7B} backbone, where Zero-shot surpasses nearly all static retrieval methods. This indicates that fixed or heuristically retrieved demonstrations often introduce noise or irrelevant patterns that disrupt the model's intrinsic reasoning process, making them more harmful than helpful.

In contrast, PICL demonstrates a consistent advantage over all baselines, particularly on challenging benchmarks. On AIME24, PICL establishes a clear lead, exceeding runner-up methods by 10.0 points on \textit{R1-Qwen-7B} and 6.6 points on \textit{R1-Llama-8B}. This consistent effectiveness across diverse backbones validates PICL’s dynamic mechanism in identifying mid-inference confusion points, proving that adaptive demonstration insertion is essential for unlocking the full potential of the reasoning models on mathematical tasks.

Finally, to provide a comprehensive evaluation, we analyze the computational efficiency of PICL compared to static baselines during reasoning process. Detailed results are provided in Appendix~\ref{sec: efficiency}.

\subsection{Ablation Studies}
To evaluate the contributions of the core components within PICL, we conducted ablation studies.

\subsubsection{Impact of Stage 1.}
To examine the contribution of Stage~1, we remove the confusion identification and summarization module and instead insert a semantically retrieved demonstration at every interruption point, regardless of whether the model is actually confused. As shown in Table~\ref{tab:abalation_study}, this variant underperforms the full PICL model, which suggests that the effectiveness of demonstrations depends on being triggered at the right moments. When demonstration insertions are applied indiscriminately, the model may receive guidance even when it does not require external support, in which case the injected examples can be distracting. Therefore, Stage~1 is crucial for determining \emph{when} to provide assistance.

\subsubsection{Impact of Stage 2.}
For the Stage~2 ablation, we retain Stage~1 to detect confusion points but disable the retrieval and re-ranking procedure in Stage~2, replacing it with random demonstration insertion once confusion is detected. As reported in Table~\ref{tab:abalation_study}, removing Stage~2 leads to a clear and consistent degradation across all benchmarks, with the largest drops on more challenging datasets (e.g., AMC23 and AIME24 on the \textit{R1-Qwen-7B} setting). These results indicate that identifying confusion is not sufficient on its own. Improvements require resolving the confusion with confusion-based selected examples. 
\begin{figure}
    \centering
    \includegraphics[width=1.0\linewidth]{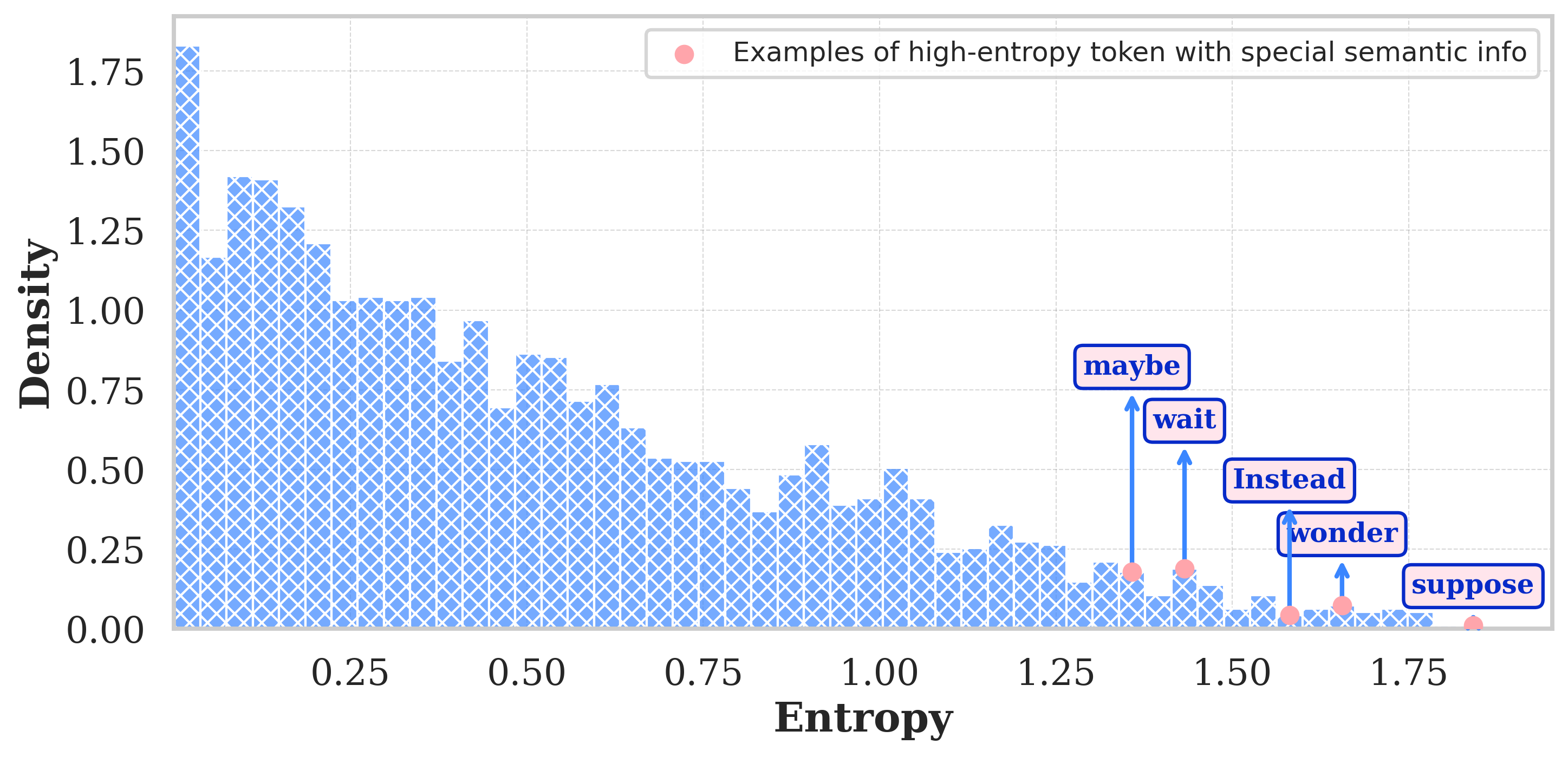}
    \caption{Density distribution of token entropy across reasoning tasks, highlighting high-entropy tokens (e.g., ``maybe'', ``wait'') that align with semantic markers of the model's uncertainty and cognitive effort.}
    \label{fig:entropy}
\end{figure}

\subsection{Analysis of Confusion Points and Entropy}
To quantitatively identify moments of uncertainty in the model's reasoning process, we employed token entropy as our primary metric. As depicted in Fig.~\ref{fig:entropy}, we analyzed the entropy of over 150,000 tokens generated across various reasoning tasks. The resulting density distribution is prominently right-skewed, indicating that the model generates the vast majority of its tokens with high certainty (low entropy). However, a critical ``long tail'' exists, composed of a small fraction of high-entropy tokens. These points represent rare but significant moments when the model faces substantial uncertainty in its decision-making.

To interpret these moments of statistical uncertainty, we examined their semantic content. Specifically, we analyzed an example set of ``interruption tokens'' that linguistically signify hesitation, reflection, or self-correction (e.g., ``maybe'', ``wait''). The analysis revealed an alignment: these semantically significant tokens consistently fall within the high-entropy region of the distribution.

This establishes an intrinsic link between the model's statistical uncertainty (high entropy) and its linguistic expression of cognitive effort. It therefore provides an empirical validation for using these interruption tokens as reliable indicators of confusion points, justifying the foundation of our intervention strategy.

\subsection{Sensitivity Analysis of PICL}
We analyze the sensitivity of PICL to two key hyperparameters: the maximum number of interventions $r$ and the number of demonstrations inserted per intervention $k$, aiming to characterize their effect on accuracy and identify robust settings.

\subsubsection{Impact of Max Interventions $r$.}
\begin{figure}
    \centering
    \includegraphics[width=1.0\linewidth]{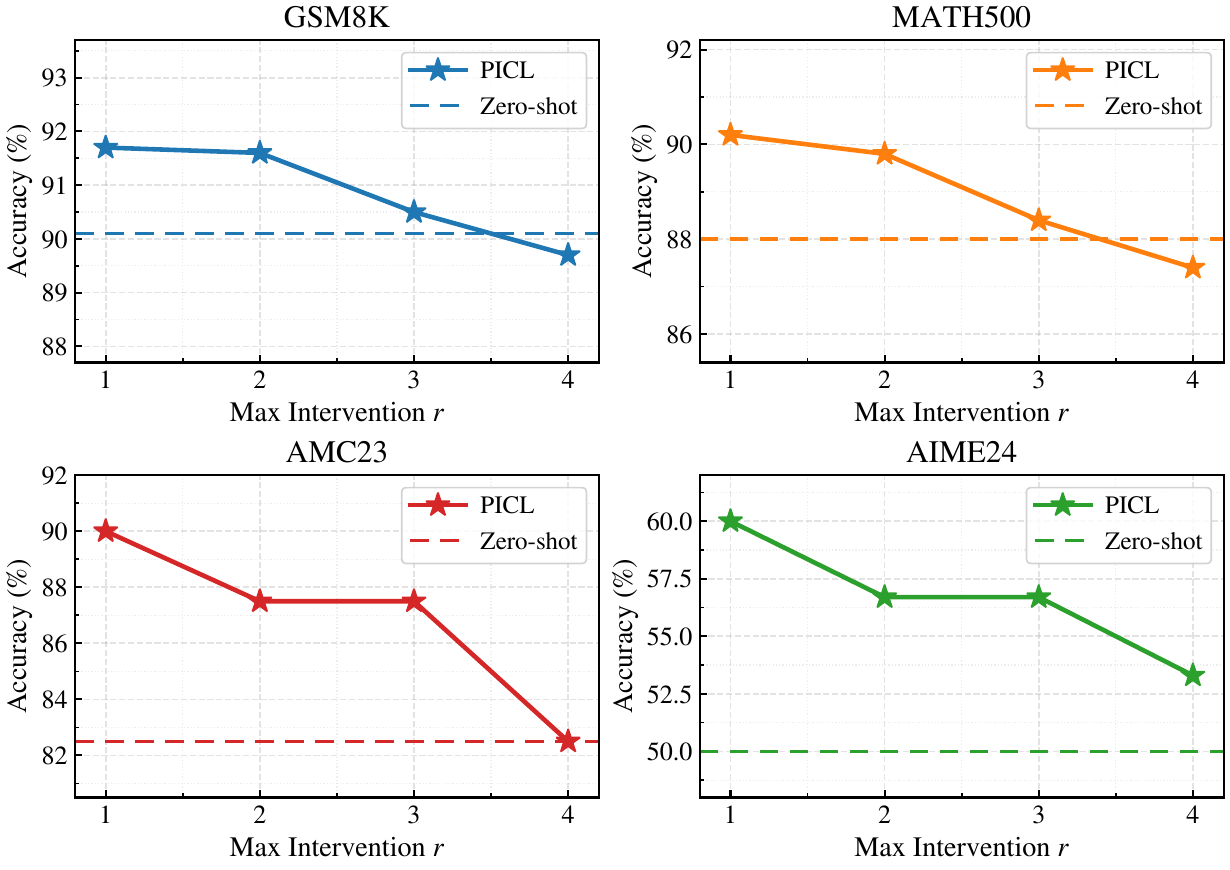}
    \caption{Performance comparison between PICL and Zero-shot with $r$ ranging from $1$ to $4$.}
    \label{fig:r_line}
\end{figure}

Fig.~\ref{fig:r_line} compares PICL with Zero-shot while varying $r$ from $1$ to $4$. Across datasets, performance is generally best at small $r$ (typically $r=1$; occasionally comparable at $r=2$) and degrades as $r$ increases. In particular, results at $r=3$ and $r=4$ are consistently lower than those at $r\le2$. This suggests that a small number of targeted interventions is beneficial, whereas frequent interventions can interrupt the model’s reasoning continuity and introduce additional noise, hurting final accuracy.

\subsubsection{Impact of Insertion Count $k$.}
\begin{figure}
    \centering
    \includegraphics[width=1.0\linewidth]{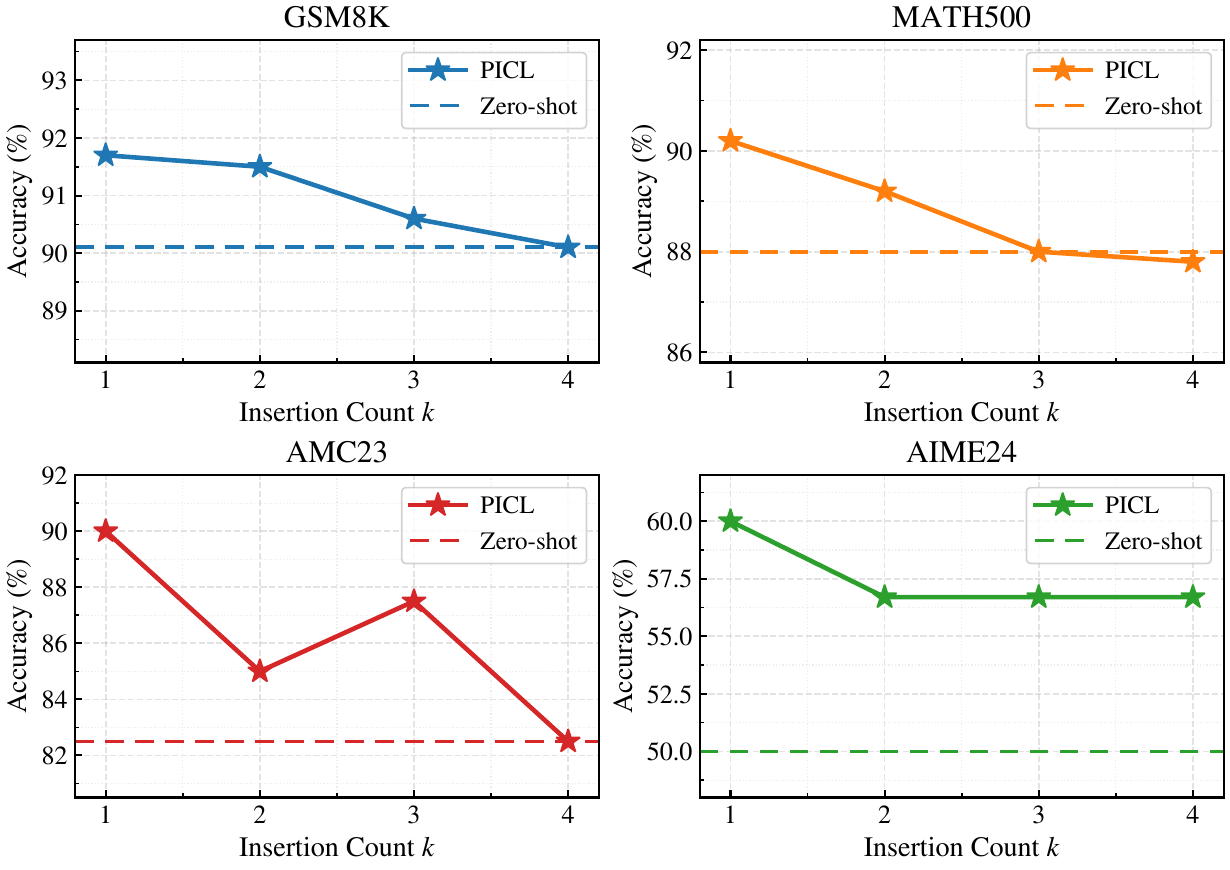}
    \caption{Performance comparison between PICL and Zero-shot with $k$ ranging from $1$ to $4$.}
    \label{fig:k_line}
\end{figure}

We further vary $k$ from $1$ to $4$ (Fig.~\ref{fig:k_line}). PICL performs best when $k$ is small, peaking at $k=1$. Notably, $k=1$ achieves the strongest results on GSM8K (91.7\%), MATH500 (90.2\%), AMC23 (90.0\%), and AIME24 (60.0\%). Increasing $k$ to $3$ or $4$ leads to a frequent drop across tasks. Overall, these results indicate that effective guidance is primarily driven by relevance rather than quantity: inserting one or two well-matched demonstrations is sufficient for course correction, while larger $k$ may introduce competing patterns that dilute the most useful signal.

\section{Conclusion}
This work addresses a key limitation of existing In-Context Learning (ICL) methods in mathematical reasoning: their static use of pre-selected demonstrations, which fail to adapt to dynamic confusion points during multi-step reasoning and often cause cascading errors. We propose Process In-Context Learning (PICL), a dynamic framework that detects confusion via semantic and entropy analysis, then retrieves and inserts relevant demonstrations to guide subsequent reasoning. Experiments show PICL outperforms baselines by mitigating mid-reasoning confusion, validating the value of adaptive demonstration insertion in complex mathematical reasoning. This work advances ICL from static to dynamic demonstration usage, with implications for other step-by-step reasoning tasks.

\section*{Limitations}
Our proposed method PICL has the following limitations. First, we validate the proposed method only on mathematical reasoning benchmarks, and it is still an open question whether the observed improvements will extend to other domains such as planning, commonsense reasoning, or more open-ended problem solving. Second, the method’s effectiveness is tied to the quality and coverage of the example pool: when the pool is sparse, domain-mismatched, or contains noisy solutions, the retrieved demonstrations may be less helpful and can sometimes be misleading. Finally, our current strategy for addressing model “confusion” relies on an external, inference-time intervention mechanism rather than internalizing this capability within the model itself, leaving internalization through training or distillation to future work.
\bibliography{main}

@article{zhang2019gap,
  title={The gap of semantic parsing: A survey on automatic math word problem solvers},
  author={Zhang, Dongxiang and Wang, Lei and Zhang, Luming and Dai, Bing Tian and Shen, Heng Tao},
  journal={IEEE transactions on pattern analysis and machine intelligence},
  volume={42},
  number={9},
  pages={2287--2305},
  year={2019},
  publisher={IEEE}
}

@article{wies2023learnability,
  title={The learnability of in-context learning},
  author={Wies, Noam and Levine, Yoav and Shashua, Amnon},
  journal={Advances in Neural Information Processing Systems},
  volume={36},
  pages={36637--36651},
  year={2023}
}

@inproceedings{min2022rethinking,
  title={Rethinking the Role of Demonstrations: What Makes In-Context Learning Work?},
  author={Min, Sewon and Lyu, Xinxi and Holtzman, Ari and Artetxe, Mikel and Lewis, Mike and Hajishirzi, Hannaneh and Zettlemoyer, Luke},
  booktitle={Proceedings of the 2022 Conference on Empirical Methods in Natural Language Processing},
  pages={11048--11064},
  year={2022}
}

@article{robertson2009probabilistic,
  title={The probabilistic relevance framework: BM25 and beyond},
  author={Robertson, Stephen and Zaragoza, Hugo and others},
  journal={Foundations and Trends{\textregistered} in Information Retrieval},
  volume={3},
  number={4},
  pages={333--389},
  year={2009},
  publisher={Now Publishers, Inc.}
}

@inproceedings{levy2022diverse,
  title={Diverse demonstrations improve in-context compositional generalization},
  author={Levy, Itay and Bogin, Ben and Berant, Jonathan},
  booktitle={Proceedings of the 61st Annual Meeting of the Association for Computational Linguistics (Volume 1: Long Papers)},
  pages={1401--1422},
  year={2023}
}

@inproceedings{liu-etal-2022-makes,
    title = "What Makes Good In-Context Examples for {GPT}-3?",
    author = "Liu, Jiachang  and
      Shen, Dinghan  and
      Zhang, Yizhe  and
      Dolan, Bill  and
      Carin, Lawrence  and
      Chen, Weizhu",
    booktitle = "Proceedings of Deep Learning Inside Out (DeeLIO 2022): The 3rd Workshop on Knowledge Extraction and Integration for Deep Learning Architectures",
    year = "2022",
    pages = "100--114",
}

@inproceedings{wu2022self,
  title={Self-adaptive in-context learning: An information compression perspective for in-context example selection and ordering},
  author={Wu, Zhiyong and Wang, Yaoxiang and Ye, Jiacheng and Kong, Lingpeng},
  booktitle={Proceedings of the 61st Annual Meeting of the Association for Computational Linguistics (Volume 1: Long Papers)},
  pages={1423--1436},
  year={2023}
}

@article{nguyen2023context,
  title={In-context example selection with influences},
  author={Nguyen, Tai and Wong, Eric},
  journal={arXiv preprint arXiv:2302.11042},
  year={2023}
}

@inproceedings{rubin2021learning,
    title = "Learning To Retrieve Prompts for In-Context Learning",
    author = "Rubin, Ohad  and
      Herzig, Jonathan  and
      Berant, Jonathan",
    editor = "Carpuat, Marine  and
      de Marneffe, Marie-Catherine  and
      Meza Ruiz, Ivan Vladimir",
    booktitle = "Proceedings of the 2022 Conference of the North American Chapter of the Association for Computational Linguistics: Human Language Technologies",
    month = jul,
    year = "2022",
    address = "Seattle, United States",
    publisher = "Association for Computational Linguistics",
    url = "https://aclanthology.org/2022.naacl-main.191/",
    doi = "10.18653/v1/2022.naacl-main.191",
    pages = "2655--2671",
}

@inproceedings{ye2023compositional,
  title={Compositional exemplars for in-context learning},
  author={Ye, Jiacheng and Wu, Zhiyong and Feng, Jiangtao and Yu, Tao and Kong, Lingpeng},
  booktitle={International Conference on Machine Learning},
  pages={39818--39833},
  year={2023},
  organization={PMLR}
}

@inproceedings{zhang2022active,
  title={Active Example Selection for In-Context Learning},
  author={Zhang, Yiming and Feng, Shi and Tan, Chenhao},
  booktitle={Proceedings of the 2022 Conference on Empirical Methods in Natural Language Processing},
  year={2022}
}

@inproceedings{wang2023learning,
  title={Learning to retrieve in-context examples for large language models},
  author={Wang, Liang and Yang, Nan and Wei, Furu},
  booktitle={Proceedings of the 18th Conference of the European Chapter of the Association for Computational Linguistics (Volume 1: Long Papers)},
  pages={1752--1767},
  year={2024}
}

@article{brown2020language,
  title={Language models are few-shot learners},
  author={Brown, Tom and Mann, Benjamin and Ryder, Nick and Subbiah, Melanie and Kaplan, Jared D and Dhariwal, Prafulla and Neelakantan, Arvind and Shyam, Pranav and Sastry, Girish and Askell, Amanda and others},
  journal={Advances in neural information processing systems},
  volume={33},
  pages={1877--1901},
  year={2020}
}

@inproceedings{devlin2019bert,
  title={Bert: Pre-training of deep bidirectional transformers for language understanding},
  author={Devlin, Jacob and Chang, Ming-Wei and Lee, Kenton and Toutanova, Kristina},
  booktitle={Proceedings of the 2019 conference of the North American chapter of the association for computational linguistics: human language technologies, volume 1 (long and short papers)},
  pages={4171--4186},
  year={2019}
}

@inproceedings{chen2024bge,
    title = "{M}3-Embedding: Multi-Linguality, Multi-Functionality, Multi-Granularity Text Embeddings Through Self-Knowledge Distillation",
    author = "Chen, Jianlyu  and
      Xiao, Shitao  and
      Zhang, Peitian  and
      Luo, Kun  and
      Lian, Defu  and
      Liu, Zheng",
    editor = "Ku, Lun-Wei  and
      Martins, Andre  and
      Srikumar, Vivek",
    booktitle = "Findings of the Association for Computational Linguistics: ACL 2024",
    month = aug,
    year = "2024",
    address = "Bangkok, Thailand",
    publisher = "Association for Computational Linguistics",
    doi = "10.18653/v1/2024.findings-acl.137",
    pages = "2318--2335",
}

@inproceedings{sorensen2022information,
  title={An information-theoretic approach to prompt engineering without ground truth labels},
  author={Sorensen, Taylor and Robinson, Joshua and Rytting, Christopher and Shaw, Alexander and Rogers, Kyle and Delorey, Alexia and Khalil, Mahmoud and Fulda, Nancy and Wingate, David},
  booktitle={Proceedings of the 60th Annual Meeting of the Association for Computational Linguistics (Volume 1: Long Papers)},
  pages={819--862},
  year={2022}
}

@inproceedings{gonen2022demystifying,
  title={Demystifying prompts in language models via perplexity estimation},
  author={Gonen, Hila and Iyer, Srini and Blevins, Terra and Smith, Noah A and Zettlemoyer, Luke},
  booktitle={Findings of the Association for Computational Linguistics: EMNLP 2023},
  pages={10136--10148},
  year={2023}
}

@inproceedings{lu2021fantastically,
  title={Fantastically ordered prompts and where to find them: Overcoming few-shot prompt order sensitivity},
  author={Lu, Yao and Bartolo, Max and Moore, Alastair and Riedel, Sebastian and Stenetorp, Pontus},
  booktitle={Proceedings of the 60th Annual Meeting of the Association for Computational Linguistics (Volume 1: Long Papers)},
  pages={8086--8098},
  year={2022}
}

@article{yao2023tree,
  title={Tree of thoughts: Deliberate problem solving with large language models},
  author={Yao, Shunyu and Yu, Dian and Zhao, Jeffrey and Shafran, Izhak and Griffiths, Tom and Cao, Yuan and Narasimhan, Karthik},
  journal={Advances in neural information processing systems},
  volume={36},
  pages={11809--11822},
  year={2023}
}

@article{chen2021evaluating,
  title={Evaluating large language models trained on code},
  author={Chen, Mark and Tworek, Jerry and Jun, Heewoo and Yuan, Qiming and Pinto, Henrique Ponde De Oliveira and Kaplan, Jared and Edwards, Harri and Burda, Yuri and Joseph, Nicholas and Brockman, Greg and others},
  journal={arXiv preprint arXiv:2107.03374},
  year={2021}
}

@inproceedings{gao2023pal,
  title={Pal: Program-aided language models},
  author={Gao, Luyu and Madaan, Aman and Zhou, Shuyan and Alon, Uri and Liu, Pengfei and Yang, Yiming and Callan, Jamie and Neubig, Graham},
  booktitle={International Conference on Machine Learning},
  pages={10764--10799},
  year={2023},
  organization={PMLR}
}

@article{lewis2020retrieval,
  title={Retrieval-augmented generation for knowledge-intensive nlp tasks},
  author={Lewis, Patrick and Perez, Ethan and Piktus, Aleksandra and Petroni, Fabio and Karpukhin, Vladimir and Goyal, Naman and K{\"u}ttler, Heinrich and Lewis, Mike and Yih, Wen-tau and Rockt{\"a}schel, Tim and others},
  journal={Advances in neural information processing systems},
  volume={33},
  pages={9459--9474},
  year={2020}
}

@article{nakano2021webgpt,
  title={Webgpt: Browser-assisted question-answering with human feedback},
  author={Nakano, Reiichiro and Hilton, Jacob and Balaji, Suchir and Wu, Jeff and Ouyang, Long and Kim, Christina and Hesse, Christopher and Jain, Shantanu and Kosaraju, Vineet and Saunders, William and others},
  journal={arXiv preprint arXiv:2112.09332},
  year={2021}
}

@article{schick2023toolformer,
  title={Toolformer: Language models can teach themselves to use tools},
  author={Schick, Timo and Dwivedi-Yu, Jane and Dess{\`\i}, Roberto and Raileanu, Roberta and Lomeli, Maria and Hambro, Eric and Zettlemoyer, Luke and Cancedda, Nicola and Scialom, Thomas},
  journal={Advances in Neural Information Processing Systems},
  volume={36},
  pages={68539--68551},
  year={2023}
}

@article{lewkowycz2022solving,
  title={Solving quantitative reasoning problems with language models},
  author={Lewkowycz, Aitor and Andreassen, Anders and Dohan, David and Dyer, Ethan and Michalewski, Henryk and Ramasesh, Vinay and Slone, Ambrose and Anil, Cem and Schlag, Imanol and Gutman-Solo, Theo and others},
  journal={Advances in neural information processing systems},
  volume={35},
  pages={3843--3857},
  year={2022}
}

@article{cobbe2021training,
  title={Training verifiers to solve math word problems},
  author={Cobbe, Karl and Kosaraju, Vineet and Bavarian, Mohammad and Chen, Mark and Jun, Heewoo and Kaiser, Lukasz and Plappert, Matthias and Tworek, Jerry and Hilton, Jacob and Nakano, Reiichiro and others},
  journal={arXiv preprint arXiv:2110.14168},
  year={2021}
}

@inproceedings{lightman2023let,
  title={Let's verify step by step},
  author={Lightman, Hunter and Kosaraju, Vineet and Burda, Yuri and Edwards, Harrison and Baker, Bowen and Lee, Teddy and Leike, Jan and Schulman, John and Sutskever, Ilya and Cobbe, Karl},
  booktitle={The Twelfth International Conference on Learning Representations},
  year={2023}
}

@inproceedings{yu2023metamath,
  title={MetaMath: Bootstrap Your Own Mathematical Questions for Large Language Models},
  author={Yu, Longhui and Jiang, Weisen and Shi, Han and YU, Jincheng and Liu, Zhengying and Zhang, Yu and Kwok, James and Li, Zhenguo and Weller, Adrian and Liu, Weiyang},
  booktitle={The Twelfth International Conference on Learning Representations},
  year={2023}
}

@article{wei2022chain,
  title={Chain-of-thought prompting elicits reasoning in large language models},
  author={Wei, Jason and Wang, Xuezhi and Schuurmans, Dale and Bosma, Maarten and Xia, Fei and Chi, Ed and Le, Quoc V and Zhou, Denny and others},
  journal={Advances in neural information processing systems},
  volume={35},
  pages={24824--24837},
  year={2022}
}

@inproceedings{guo2024makes,
  title={What makes a good order of examples in in-context learning},
  author={Guo, Qi and Wang, Leiyu and Wang, Yidong and Ye, Wei and Zhang, Shikun},
  booktitle={Findings of the Association for Computational Linguistics: ACL 2024},
  pages={14892--14904},
  year={2024}
}

@inproceedings{liu2021makes,
  title={What makes good in-context examples for GPT-3?},
  author={Liu, Jiachang and Shen, Dinghan and Zhang, Yizhe and Dolan, William B and Carin, Lawrence and Chen, Weizhu},
  booktitle={Proceedings of Deep Learning Inside Out (DeeLIO 2022): The 3rd workshop on knowledge extraction and integration for deep learning architectures},
  pages={100--114},
  year={2022}
}

@article{zheng2025curse,
  title={The curse of cot: On the limitations of chain-of-thought in in-context learning},
  author={Zheng, Tianshi and Chen, Yixiang and Li, Chengxi and Li, Chunyang and Zong, Qing and Shi, Haochen and Xu, Baixuan and Song, Yangqiu and Wong, Ginny Y and See, Simon},
  journal={arXiv preprint arXiv:2504.05081},
  year={2025}
}

@article{tang2025refcritic,
  title={RefCritic: Training Long Chain-of-Thought Critic Models with Refinement Feedback},
  author={Tang, Qiaoyu and Xiang, Hao and Yu, Le and Yu, Bowen and Lin, Hongyu and Lu, Yaojie and Han, Xianpei and Sun, Le and Lin, Junyang},
  journal={arXiv preprint arXiv:2507.15024},
  year={2025}
}

@article{cheng2025revisiting,
  title={Revisiting Chain-of-Thought Prompting: Zero-shot Can Be Stronger than Few-shot},
  author={Cheng, Xiang and Pan, Chengyan and Zhao, Minjun and Li, Deyang and Liu, Fangchao and Zhang, Xinyu and Zhang, Xiao and Liu, Yong},
  journal={arXiv preprint arXiv:2506.14641},
  year={2025}
}

@article{wang2025beyond,
  title={Beyond the 80/20 rule: High-entropy minority tokens drive effective reinforcement learning for llm reasoning},
  author={Wang, Shenzhi and Yu, Le and Gao, Chang and Zheng, Chujie and Liu, Shixuan and Lu, Rui and Dang, Kai and Chen, Xionghui and Yang, Jianxin and Zhang, Zhenru and others},
  journal={arXiv preprint arXiv:2506.01939},
  year={2025}
}

@inproceedings{dong2022survey,
  title={A survey on in-context learning},
  author={Dong, Qingxiu and Li, Lei and Dai, Damai and Zheng, Ce and Ma, Jingyuan and Li, Rui and Xia, Heming and Xu, Jingjing and Wu, Zhiyong and Chang, Baobao and others},
  booktitle={Proceedings of the 2024 conference on empirical methods in natural language processing},
  pages={1107--1128},
  year={2024}
}

@inproceedings{liu2024makes,
  title={What Makes In-context Learning Effective for Mathematical Reasoning},
  author={Liu, Jiayu and Huang, Zhenya and Wang, Chaokun and Huang, Xunpeng and Zhai, ChengXiang and Chen, Enhong},
  booktitle={Forty-second International Conference on Machine Learning},
  year={2025}
}

@article{gsm8k,
  title={Training Verifiers to Solve Math Word Problems},
  author={Cobbe, Karl and Kosaraju, Vineet and Bavarian, Mohammad and Chen, Mark and Jun, Heewoo and Kaiser, Lukasz and Plappert, Matthias and Tworek, Jerry and Hilton, Jacob and Nakano, Reiichiro and Hesse, Christopher and Schulman, John},
  journal={arXiv preprint arXiv:2110.14168},
  year={2021}
}

@article{math500,
  title={Let's Verify Step by Step},
  author={Lightman, Hunter and Kosaraju, Vineet and Burda, Yura and Edwards, Harri and Baker, Bowen and Lee, Teddy and Leike, Jan and Schulman, John and Sutskever, Ilya and Cobbe, Karl},
  journal={CoRR},
  year={2023}
}

@article{mmlu,
  title={Measuring Massive Multitask Language Understanding},
  author={Dan Hendrycks and Collin Burns and Steven Basart and Andy Zou and Mantas Mazeika and Dawn Song and Jacob Steinhardt},
  journal={Proceedings of the International Conference on Learning Representations (ICLR)},
  year={2021}
}

@article{mmlu_pro,
  title={Mmlu-pro: A more robust and challenging multi-task language understanding benchmark},
  author={Wang, Yubo and Ma, Xueguang and Zhang, Ge and Ni, Yuansheng and Chandra, Abhranil and Guo, Shiguang and Ren, Weiming and Arulraj, Aaran and He, Xuan and Jiang, Ziyan and others},
  journal={Advances in Neural Information Processing Systems},
  volume={37},
  pages={95266--95290},
  year={2024}
}

@inproceedings{bgereranker,
  title={M3-embedding: Multi-linguality, multi-functionality, multi-granularity text embeddings through self-knowledge distillation},
  author={Chen, Jianlyu and Xiao, Shitao and Zhang, Peitian and Luo, Kun and Lian, Defu and Liu, Zheng},
  booktitle={Findings of the Association for Computational Linguistics ACL 2024},
  pages={2318--2335},
  year={2024}
}

@misc{deepseek,
      title={DeepSeek-R1: Incentivizing Reasoning Capability in LLMs via Reinforcement Learning}, 
      author={DeepSeek-AI},
      year={2025},
      eprint={2501.12948},
      archivePrefix={arXiv},
      primaryClass={cs.CL},
}

@inproceedings{AIME,
      author={MAA},
      year={2024},
      title = {American Invitational Mathematics Examination - AIME},
      booktitle = {American Invitational Mathematics Examination - AIME},
      url = {https://maa.org/math-competitions/american-invitational-mathematics-examination-aime}
}

@inproceedings{AMC23,
      author={MAA},
      yaer={2023},
      title = {American Mathematics Competitions},
      booktitle = {American Mathematics Competitions},
      year = {2023}
}

@article{webthinker,
  title={Webthinker: Empowering large reasoning models with deep research capability},
  author={Li, Xiaoxi and Jin, Jiajie and Dong, Guanting and Qian, Hongjin and Wu, Yongkang and Wen, Ji-Rong and Zhu, Yutao and Dou, Zhicheng},
  journal={arXiv preprint arXiv:2504.21776},
  year={2025}
}

\newpage
\appendix
\section{Case Study}\label{sec:case_study}

\begin{figure}[h]
    \centering
    \includegraphics[width=1.0\linewidth]{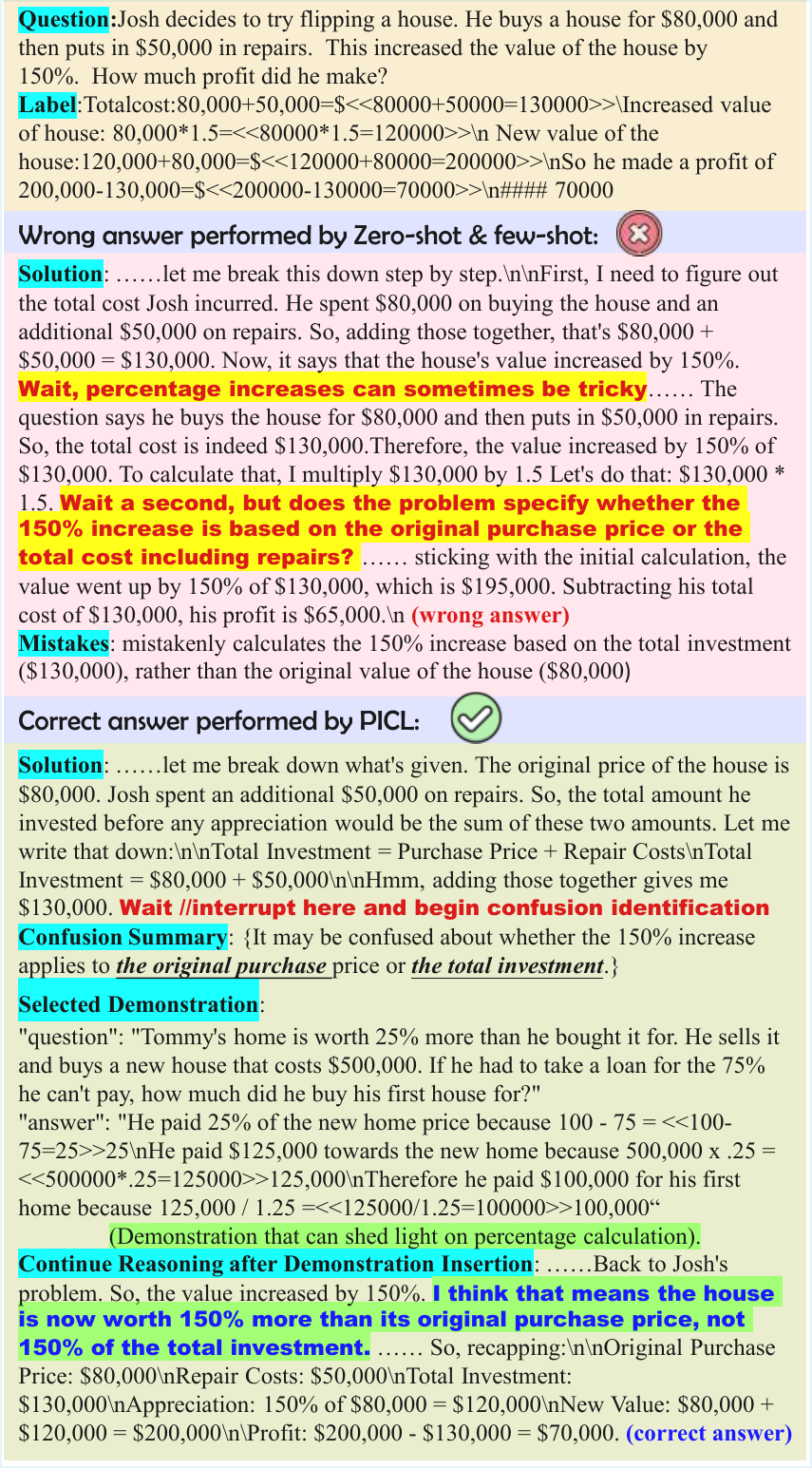}
    \caption{A case study comparing reasoning processes.}
    \label{fig:case_study}
\end{figure}

To demonstrate how PICL mitigates ambiguity in mathematical reasoning, we analyze a house-flipping profit problem, shown in Fig.~\ref{fig:case_study}. In zero-shot and static few-shot settings, the model’s reasoning hinges on an ambiguous reference point: whether the stated 150\% appreciation is applied to the purchase price or to the total expenditure including repairs (highlighted in yellow). This mis-specified baseline causes the model to follow an incorrect computation path and yield an erroneous profit.

In contrast, PICL detects this uncertainty as a confusion point and retrieves a closely matched demonstration that operationalizes percentage increases with respect to the original value. Injecting this demonstration at inference time resolves the ambiguity, leading the model to ground the 150\% increase on the purchase price and compute the correct profit. This case study suggests that PICL can localize mid-inference ambiguities and provide targeted corrective context to recover valid reasoning trajectories.





\section{Generalization to Multiple-Choice Benchmarks}\label{sec:mc_generalization}
\begin{table}[t]
\centering
\begin{tabular}{lcc}
\hline
Method & MMLU & MMLU\_Pro \\ \hline
Zero-shot           & {\underline{80.0}} & {\underline{55.2}} \\
Random              & 69.0               & 45.4               \\
Best-validate       & 76.0               & 50.3               \\
Similarity          & 70.0               & 41.9               \\
BM25                & 64.0               & 42.3               \\
BGEM3               & 73.0               & 42.6               \\
Perplexity          & 68.0               & 44.9               \\
Influence           & 57.0               & 45.9               \\
LMS3                & 71.0               & 43.5               \\
\textbf{PICL (Ours)} & \textbf{85.0}      & \textbf{57.9}      \\ \hline
\end{tabular}
\caption{
Accuracy (\%) comparasion on MMLU and MMLU\_Pro, two multiple-choice reasoning benchmarks.
All few-shot baselines utilize one-shot reasoning. For PICL, parameters are set
to k = 1 and r = 1. The best and runner-up methods are highlighted in bold and underlined, respectively
}
\label{tab:mc_generalization}
\end{table}

Although PICL is introduced for open-ended mathematical reasoning with short-answer supervision, we further test its robustness under a format shift to multiple-choice evaluation. Specifically, we evaluate on MMLU~\cite{mmlu} and MMLU\_Pro~\cite{mmlu_pro}, whose tasks and output space (option selection) differ substantially from the short-answer benchmarks used in our main experiments. We carry out this experiment on Deepseek-R1-Distilled-Qwen-7B.

Table~\ref{tab:mc_generalization} shows that PICL achieves the best performance on both datasets, improving over zero-shot by +5.0 on MMLU (80.0$\rightarrow$85.0) and +2.7 on MMLU\_Pro (55.2$\rightarrow$57.9). Notably, several retrieval- or selection-based baselines underperform zero-shot, suggesting that statically chosen demonstrations can induce negative transfer under multiple-choice prompting. In contrast, PICL remains consistently effective, indicating that confusion-aware retrieval and dynamic demonstration insertion transfer beyond open-ended generation and provide robust gains under choice-based reasoning.

We further analyze the sensitivity of PICL on multiple-choice benchmarks with respect to the same two hyperparameters: the maximum number of interventions $r$ and the number of demonstrations inserted per intervention $k$.

\begin{figure}
    \centering
    \includegraphics[width=1.0\linewidth]{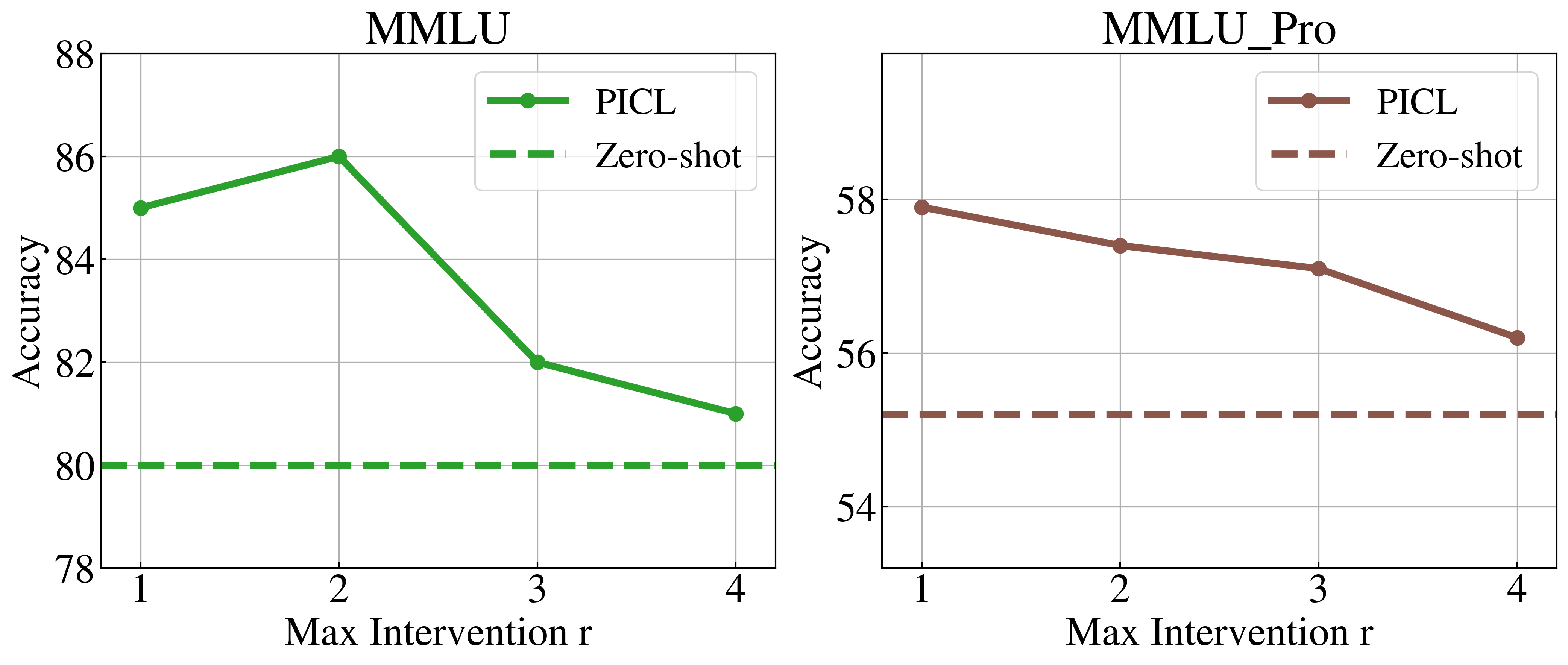}
    \caption{Performance comparison between PICL and
Zero-shot with r ranging from 1 to 4 on multiple-choice benchmarks.}
    \label{fig:r_line_mmlu}
\end{figure}

Fig.~\ref{fig:r_line_mmlu} reports the results when varying $r$ from $1$ to $4$ on MMLU and MMLU\_Pro. We observe that PICL achieves its strongest performance under a small number of interventions (typically $r\in\{1,2\}$), while increasing $r$ beyond $2$ leads to a clear degradation. This trend mirrors our findings on open-ended reasoning tasks: a small number of well-timed interventions can effectively correct the model’s trajectory, whereas frequent interruptions may disrupt the continuity of choice-based reasoning and introduce additional contextual noise, ultimately hurting accuracy.

\begin{figure}
    \centering
    \includegraphics[width=1.0\linewidth]{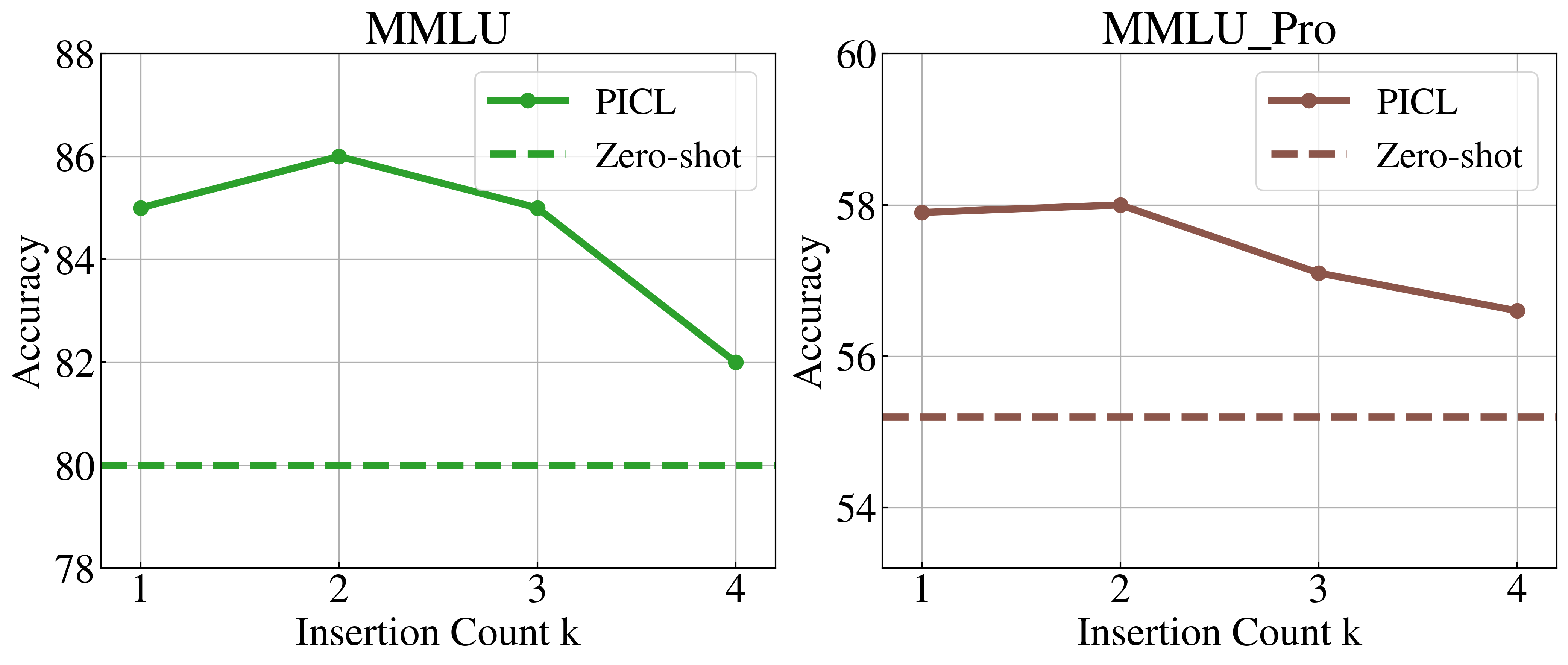}
    \caption{Performance comparison between PICL and
Zero-shot with k ranging from 1 to 4 on multiple-choice benchmarks.}
    \label{fig:k_line_mmlu}
\end{figure}

We also vary $k$ from $1$ to $4$ in Fig.~\ref{fig:k_line_mmlu}. PICL is most robust when $k$ is small (typically $k\in\{1,2\}$), and larger $k$ does not yield further gains and can slightly reduce performance. Overall, these results suggest that even under multiple-choice prompting, PICL benefits primarily from relevance rather than quantity: inserting one (or at most two) highly matched demonstrations is sufficient to resolve confusion, while inserting more examples may introduce competing patterns that dilute the most useful signal.

\section{Efficiency Analysis}\label{sec: efficiency}

\begin{figure}[h]
    \centering
    \includegraphics[width=1.0\linewidth]{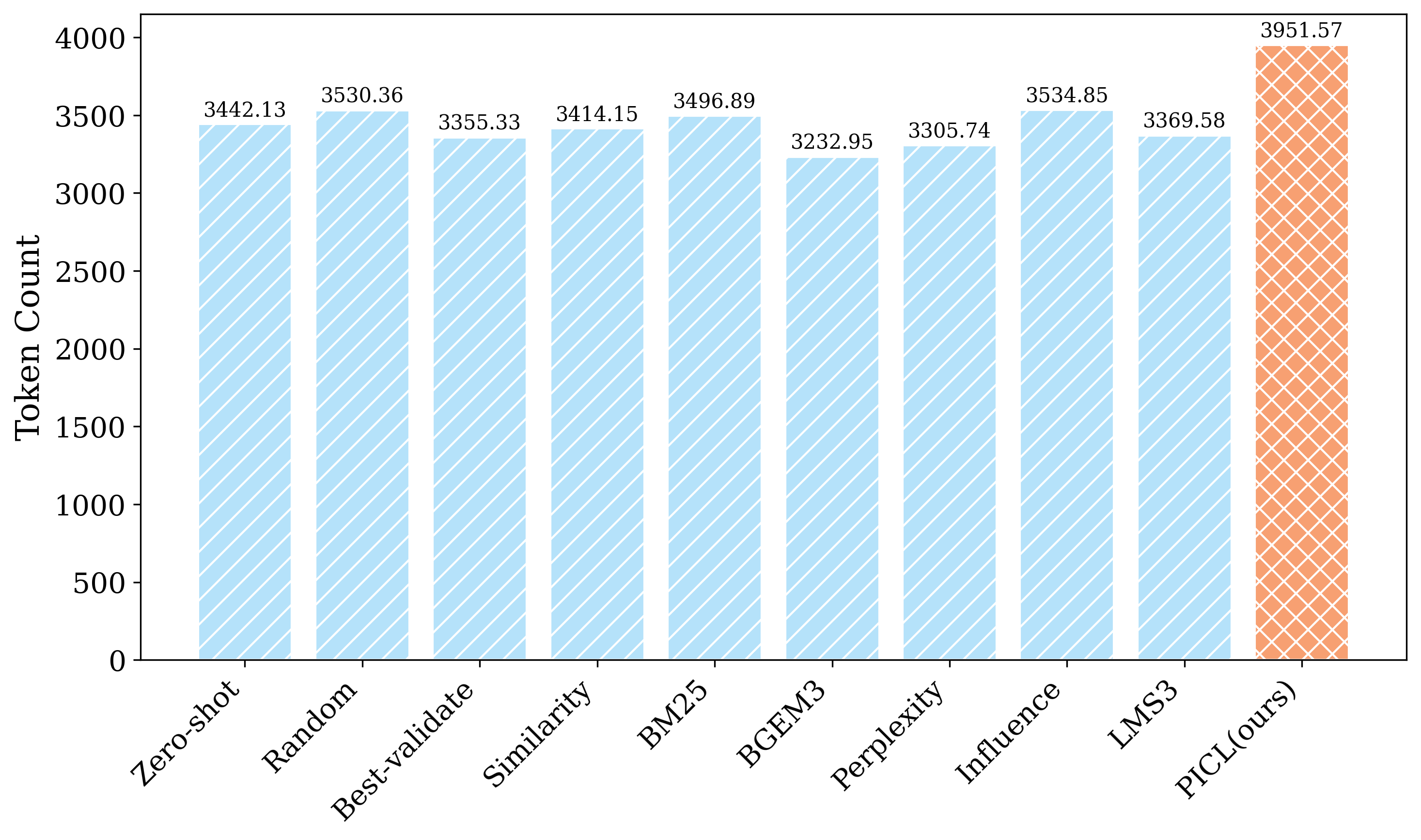}
    \caption{Average token used per question for reasoning in different methods.}
    \label{fig:token usage}
\end{figure}

To quantify the efficiency of the proposed method, we conducted a statistical analysis of the inference length using DeepSeek-R1-Distilled-Qwen-7B. We aggregated the number of tokens generated for the reasoning chain of each query across all evaluated datasets. As shown in Fig.~\ref{fig:token usage}, the average token consumption for PICL is 3,951, compared to an average of 3,409 for the baseline approaches. This comparison demonstrates that the absolute token usage of PICL remains within a reasonable range, ensuring that the enhanced reasoning capability does not come at the expense of excessive computational latency.

\section{Prompt Design}\label{sec: prompt}
\begin{figure}
    \centering
    \includegraphics[width=0.9\linewidth]{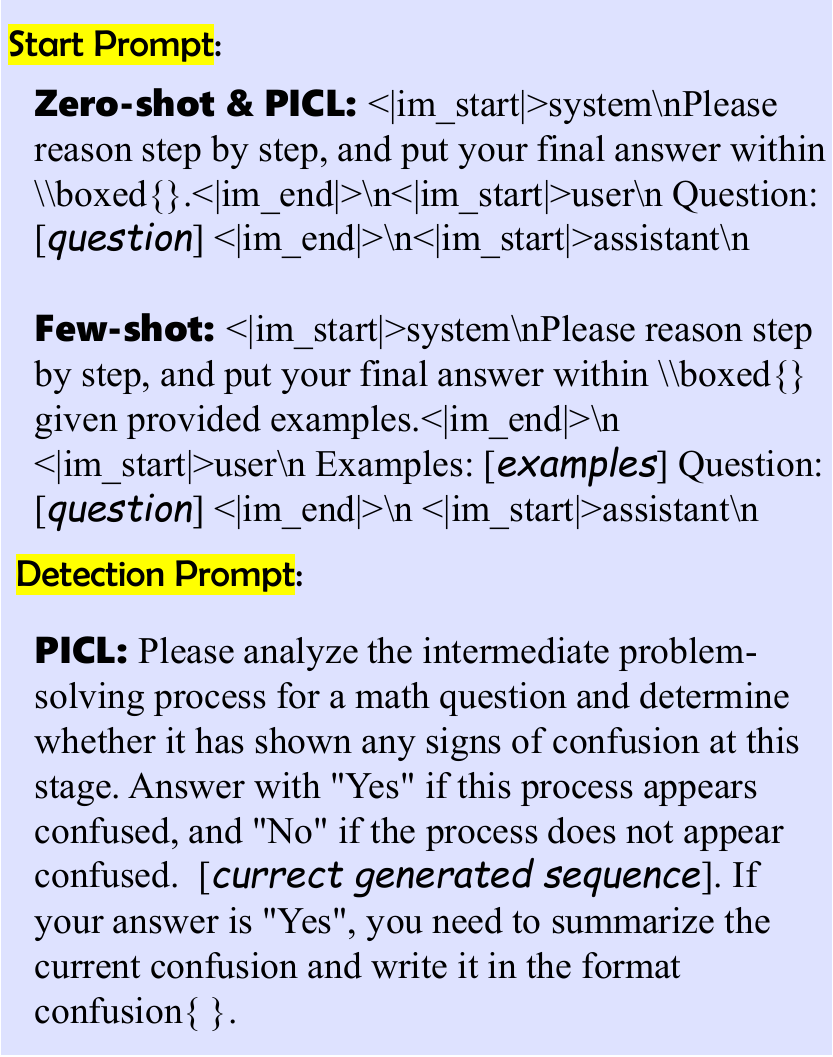}
    \caption{An overview of prompt used in this work.}
    \label{fig:prompt}
\end{figure}

To implement the aforementioned stages, PICL employs specific prompt designs tailored to different reasoning scenarios and the confusion detection task, which is shown in Fig.~\ref{fig:prompt}. For zero-shot and PICL settings, the prompt structure is designed to guide the model in step-by-step reasoning, with a requirement to enclose the final answer within \\boxed\{\}. This ensures a standardized output format for consistent evaluation. In few-shot scenarios, the prompt is augmented with provided examples, prompting the model to leverage these examples for reasoning while still adhering to the final answer formatting requirement.

Notably, the Detection Prompt plays a critical role in the Confusion Identification and Summarization stage. It instructs the model to analyze the intermediate problem-solving process, determine if there are signs of confusion, and answer with "Yes" or "No". If confusion is identified ("Yes"), the prompt further requires a concise summary of the confusion in a specified format confusion\{\}, which directly contributes to generating the confusion summary \(\mathcal{C}\) as formalized in the stage description. These prompt designs collectively facilitate the effective execution of PICL's core mechanisms, ensuring accurate confusion detection and contextually relevant demonstration integration.

\end{document}